%% file: main.tex
\documentclass[11pt, a4paper]{article}
\usepackage[hyperref]{acl2020}
\usepackage{times}
\usepackage{latexsym}

\usepackage{CJKutf8}
\newcommand{\zh}[1]{\begin{CJK}{UTF8}{gbsn}#1\end{CJK}}
\usepackage{musicography}

\input{preamble}

\aclfinalcopy 

\title{ON-TRAC Consortium for End-to-End and Simultaneous Speech Translation Challenge Tasks at IWSLT 2020}

\author{Maha Elbayad$^{1,4}$\Thanks{Equal contribution.}, 
    Ha Nguyen$^{1,2}$\footnotemark[1], 
    Fethi Bougares$^3$, 
    Natalia Tomashenko$^2$,\\
    \bf
    Antoine Caubri{\`e}re$^3$,
    Benjamin Lecouteux$^1$,
    Yannick Est{\`e}ve$^2$,
    Laurent Besacier$^1$\\
    $^1$LIG - Universit{\'e} Grenoble Alpes, France  \\
    $^2$LIA - Avignon Universit{\'e}, France \\
    $^3$LIUM - Le Mans Universit{\'e}, France\\
    $^4$Inria - Grenoble, France
}

\begin{document}

\maketitle
\begin{abstract}
This paper describes the ON-TRAC Consortium translation systems developed for two challenge tracks featured in the Evaluation Campaign of IWSLT 2020, 
offline speech translation
and
simultaneous speech translation.
ON-TRAC Consortium is composed of researchers from three French academic laboratories: 
LIA (Avignon Universit\'e),
LIG (Universit\'e Grenoble Alpes), 
and LIUM (Le Mans Universit\'e).
Attention-based encoder-decoder models, trained end-to-end, were used for our submissions to the offline speech translation track. 
Our contributions focused on data augmentation and ensembling of multiple models. 
In the simultaneous speech translation track, we build on Transformer-based \waitk models 
for the text-to-text subtask.
For speech-to-text simultaneous translation, we attach a \waitk MT system to a hybrid ASR system. 
We propose an algorithm to control the latency of the ASR+MT cascade and
achieve a good latency-quality trade-off on both subtasks.
\end{abstract}

\input{intro}

\input{offline}
\input{online}

\section{Conclusion}
\label{sec:conclusion}
This paper described the ON-TRAC consortium submission to the IWSLT 2020 shared task. In the continuity of our 2019 participation, we have submitted several end-to-end systems to the offline speech translation track. A significant part of our efforts was also dedicated to the new simultaneous translation track: we improved \waitk models with unidirectional encoders and multi-path training and cascaded them with a strong ASR system. Future work will be dedicated to simultaneous speech translation using end-to-end models.

\section{Acknowledgements}

This work was funded by the French Research Agency (ANR) through the ON-TRAC project under contract number ANR-18-CE23-0021.

\bibliographystyle{acl_natbib}
\bibliography{refs}
\end{document}

%% file: preamble.tex
\usepackage{microtype}
\usepackage[T1]{fontenc}
\usepackage{placeins}
\usepackage{latexsym}
\usepackage{graphicx}
\graphicspath{{figures/}}

\usepackage{subcaption}
\usepackage{booktabs}
\usepackage{makecell}
\usepackage{float}
\graphicspath{{figures/}}

\def\mypar#1{\vspace{2mm}{\noindent\bfseries #1.}\hspace{1mm}} 

\usepackage{mathtools}
\usepackage{amssymb} 
\usepackage{mathrsfs} 
\usepackage{bbm}  
\usepackage{algorithm}
\usepackage{algpseudocode}

\algnewcommand{\IIf}[1]{\State\algorithmicif\ #1\ \algorithmicthen}
\algnewcommand{\EndIIf}{\unskip\ \algorithmicend\ \algorithmicif}
\renewcommand{\algorithmicrequire}{\textbf{Input:}~}
\renewcommand{\algorithmicensure}{\textbf{Output:}~}
\algnewcommand{\LineComment}[1]{\State \(\triangleright\) #1}
\algnewcommand{\IfThenElse}[3]{
  \State \algorithmicif\ #1\ \algorithmicthen\ #2\ \algorithmicelse\ #3}
  
 \algnewcommand{\IfThen}[2]{
  \State \algorithmicif\ #1\ \algorithmicthen\ #2}
  
\newcommand{\pluseq}{\mathrel{+}=}

\usepackage{nameref}  
\def\alg#1{Algorithm~\ref{alg:#1}}
\def\fig#1{Figure~\ref{fig:#1}}

\def\tab#1{Table~\ref{tab:#1}}
\def\sect#1{\textsection\ref{sec:#1}}

\usepackage{xspace}  
\usepackage{enumitem}

\newcommand{\E}[2]{\mathbb{E}_{#1}\!\left[#2\right]}

\newcommand{\waitk}{\mbox{\emph{wait-$k$}~}}
\newcommand{\y}{\boldsymbol y}
\newcommand{\x}{\boldsymbol x}
\newcommand{\xasr}{\boldsymbol x^\text{asr}}
\newcommand{\z}{\boldsymbol z}
\newcommand{\cond}{\,|\,}
\newcommand{\zw}{z^{\text{$k$}}}
\newcommand{\Zw}[1]{\boldsymbol z^{\text{$#1$}}}

\newcommand{\Zm}{\textrm{Z}}
\newcommand{\ra}{\textrm{READ}}
\newcommand{\wa}{\textrm{WRITE}}

\newcommand{\eos}{{<}\!{/}\!s{>}}
\newcommand{\bos}{{<}s{>}}

\newcommand{\lx}{{|\boldsymbol x|}}
\newcommand{\ly}{{|\boldsymbol y|}}
\newcommand{\ktr}{{k_\text{train}}}
\newcommand{\kev}{{k_\text{eval}}}

\makeatletter
\DeclareRobustCommand\onedot{\futurelet\@let@token\@onedot}
\def\@onedot{\ifx\@let@token.\else.\null\fi\xspace}

\def\eg{\emph{e.g}\onedot} 
\def\ie{\emph{i.e}\onedot}

\def\wrt{w.r.t\onedot} 

\makeatother

\definecolor{p1blue}{RGB}{0, 90, 200}      
\definecolor{p1red}{RGB}{170, 10, 60}      
\definecolor{p1green}{RGB}{10, 155, 75}    
\definecolor{p1yellow}{RGB}{234, 214, 68}  
\definecolor{p1orange}{RGB}{255, 130, 95}  
\definecolor{p1purple}{RGB}{130, 20, 160}  
\definecolor{p1azure}{RGB}{0, 160, 250}    
\definecolor{padgray}{rgb}{.7,.7,.7}      
\definecolor{dimgray}{rgb}{.35,.35,.35}   
\definecolor{darkgray}{rgb}{.20,.20,.20}

\usepackage{tikz, pgfplots}
\usepackage{pgffor, pgfmath}

\pgfplotsset{
    compat=1.14,
    grid style={darkgray},
    minor grid style={dimgray!20},
    major grid style={dimgray!20},
    axis line style = { darkgray }, 
    every axis plot/.append style={line width=1.5pt, mark options=solid, mark size=4pt},
    legend style={draw = darkgray, rounded corners=0pt, fill = white, font=\Large},
    tick style ={color = dimgray!30 },
    tick label style={font=\normalsize},
    label style={font=\normalsize},
}

%% file: intro.tex
\section{Introduction}\label{sec:intro}

While cascaded speech-to-text translation~(AST) systems (combining source language speech recognition~(ASR) and source-to-target text translation~(MT)) remain state-of-the-art, recent works have attempted to build end-to-end AST with very encouraging results~\cite{berard-nips2016,weiss2017sequence,berard:hal-01709586,DBLP:journals/corr/abs-1811-02050, DBLP:journals/corr/abs-1904-07209}. This year, IWSLT 2020 \textit{offline translation track} attempts to evaluate if end-to-end AST will close the gap with cascaded AST for the English-to-German language pair.

Another increasingly popular topic is simultaneous (online) machine translation which consists in generating an output hypothesis before the entire input sequence is available. 
To deal with this low latency constraint, several strategies were proposed for neural machine translation with input text~\cite{Ma19acl, Arivazhagan19acl, Ma20iclr}. 
Only a few works investigated low latency neural speech translation~\cite{DBLP:journals/corr/abs-1808-00491}.
This year, IWSLT 2020 \textit{simultaneous translation track} attempts to stimulate research on this challenging task.
This paper describes the ON-TRAC consortium automatic speech translation~(AST) systems for the IWSLT 2020 Shared Task~\cite{iwslt:2020}. ON-TRAC Consortium is composed of researchers from three French academic laboratories: LIA~(Avignon Universit\'e), LIG~(Universit\'e Grenoble Alpes), and LIUM~(Le Mans Universit\'e).

We participated in: 
\begin{itemize}
    \item IWSLT 2020 \textit{offline translation track} with end-to-end models for the English-German language pair, 
     \item IWSLT 2020 \textit{simultaneous translation track} with a cascade of an ASR system trained using Kaldi~\cite{Povey11thekaldi} and an online MT system with \emph{wait-$k$} policies~\cite{Dalvi18naacl, Ma19acl}.
\end{itemize}
    
This paper goes as follows: we review the systems built for the offline speech translation track in \sect{offline}. Then, we present our approaches to the simultaneous track for both text-to-text and speech-to-text subtasks in \sect{simultaneous}. We ultimately conclude this work in \sect{conclusion}.

%% file: offline.tex
\section{Offline Speech translation Track}\label{sec:offline}
In this work, we developed several end-to-end 
speech translation systems, using a similar architecture as last year \cite{nguyen2019ontrac} and adapting it for translating English speech into German text (En-De).
All the systems were developed using the ESPnet~\cite{watanabe2018espnet} end-to-end speech processing toolkit.

\subsection{Data and pre-processing} 
\mypar{Data}
We relied on MuST-C~\cite{mustc19} English-to-German (hereafter called MuST-C original), and Europarl~\cite{europarlst} English-to-German as our main corpora. Besides, we automatically translated (into German) the English transcription of MuST-C and How2~\cite{sanabria2018how2} in order to augment  training data. This resulted in two synthetic corpora, which are called \textit{MuST-C synthetic} and \textit{How2 synthetic} respectively. The statistics of these corpora, along with
the provided evaluation data, can be found in
Table~\ref{table:datastatistics}. We experimented with different ways of combining those corpora. The details of these experiments are presented later in this section.

\begin{table}
 \centering
 \begin{tabular}{| c | c | c | c | c |} 
  \hline
  \textbf{\thead{Name}} & \textbf{\thead{\#segments}} &  \textbf{\thead{Total length\\(in hours)}} \\ 
  \hline
  \hline
  MuST-C train & 229.703 & 400 \\ 
  \hline
  MuST-C dev & 1.423 & 2.5 \\
  \hline
  MuST-C tst-COMMON & 2.641 & 4.1 \\
  \hline
  MuST-C tst-HE & 600 & 1.2 \\
  \hline
  \hline
  Europarl train & 32.628 & 77 \\ 
  \hline
  Europarl dev & 1.320 & 3.1 \\ 
  \hline
  \hline
  How2 synthetic & 176.564 & 285.5 \\
  \hline
  \hline
  tst2019 & 2.813 & 5.1 \\ 
  \hline
  tst2020 & 2.263 & 4.1 \\ 
  \hline
 \end{tabular}
 \caption{Statistics of training and evaluation data. The statistics of tst2019 and tst2020 are measured on the segmented version provided by IWSLT2020 organizers.}
 \label{table:datastatistics}
\end{table}

\mypar{Speech features and data augmentation}
80-dimensional Mel filter-bank features, concatenated with 3-dimensional pitch features\footnote{Pitch-features are computed using the Kaldi toolkit~\cite{Povey11thekaldi} and consist of the following
values~\cite{ghahremani2014pitch}: (1) probability of voicing (POV-feature), (2) pitch-feature and (3) delta-pitch feature. For details, see \url{http://kaldi-asr.org/doc/process-kaldi-pitch-feats_8cc.html} }
are extracted from windows of $25ms$ with a frame shift of $10ms$. We computed mean and variance normalization on these raw features of the training set, then applied it on all the data. 
Beside speed perturbation with factors of 0.9, 1.0, and 1.1, SpecAugment~\cite{park2019specaugment} is applied to the training data~\cite{ko2015perturb}. All three SpecAugment methods were used, including time warping ($W=5$), frequency masking ($F=30$), and time masking ($T=40$).

\mypar{Text preprocessing} 
The same as last year, we normalize punctuation, and tokenize all the German text using Moses.\footnote{\url{http://www.statmt.org/moses/}}
Texts are case-sensitive and contain punctuation. Moreover, the texts of the MuST-C corpus contain multiple non speech events (\textit{i.e} \textit{'Laughter'}, \textit{'Applause'} etc.). All these marks are removed from the texts before training our models. 
This results in a vocabulary of 201 characters. We find that some of these characters should not appear in the German text, for example, \musEighth, \zh{你, 葱, 送}, etc. Therefore, we manually exclude them from the vocabulary. In the end, we settle with an output vocabulary of 182 characters.

\begin{figure}
    \centering
    \includegraphics[scale=0.4]{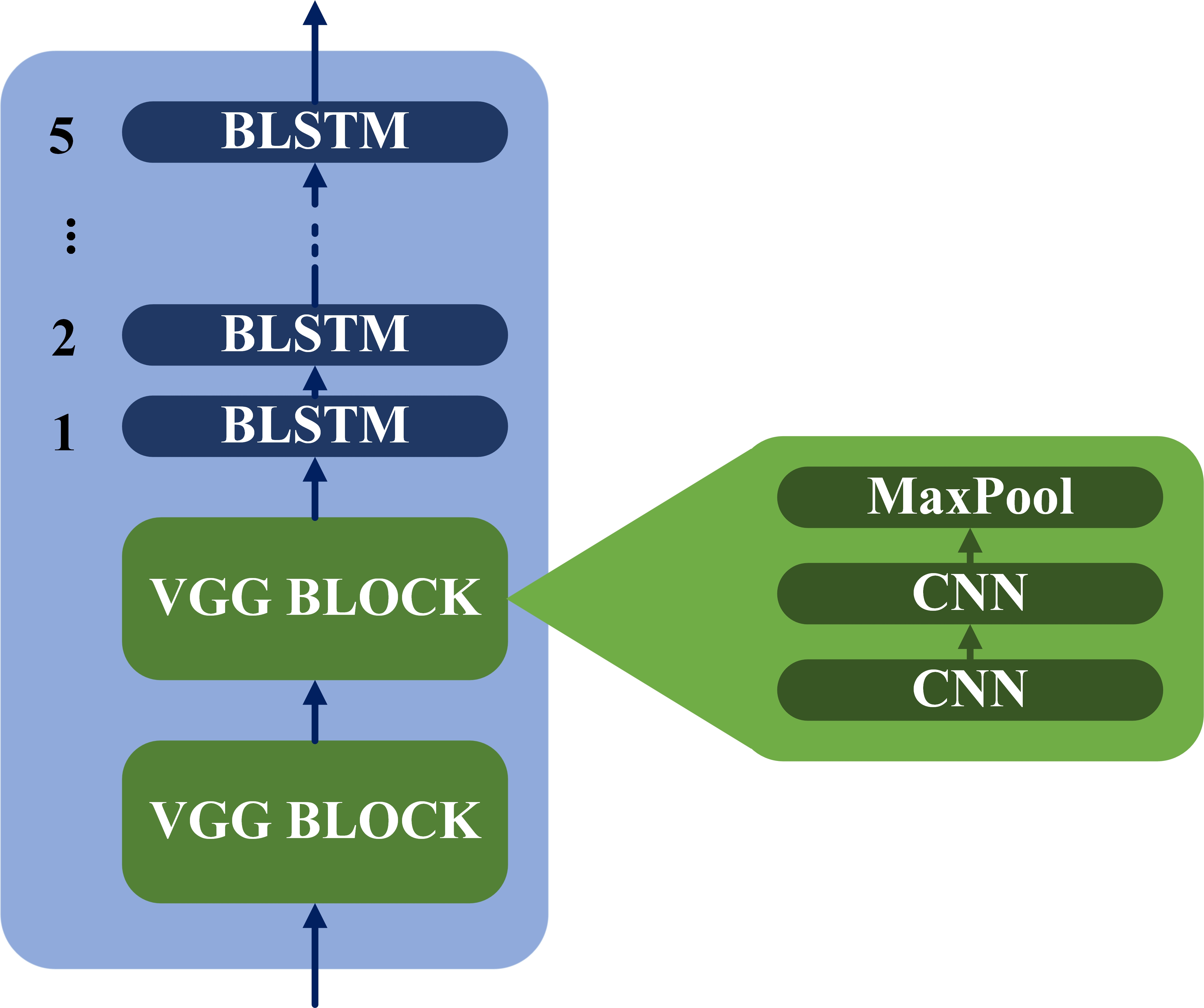}
    \centering
    \caption{Architecture of the speech encoder: a stack of two VGG blocks followed by 5 BLSTM layers.}
    \label{fig:1}
\end{figure}

\begin{table*}
 \centering
 \begin{tabular}{| c | l | c | c | c | c |} 
  \hline
  \textbf{\thead{No.}} & \textbf{\thead{Experiment}} & \textbf{\thead{MuST-C \\ tst-COMMON}} & \textbf{\thead{MuST-C \\ tst-HE}} & \textbf{tst2015 (iwslt seg)} & \textbf{tst2015 (ASR seg)} \\ 
  \hline
  \hline
  1 & \thead{MuST-C original +\\EuroParl} & 20.18 & 19.82 & 12.59 & 14.85 \\ 
  \hline
  2 & \thead{MuST-C original +\\Europarl +\\How2 synthetic} & 20.51 & 20.10 & 12.10 & 13.66 \\ 
  \hline
  3* & \thead{MuST-C original +\\Europarl +\\How2 synthetic} & 23.55 & 22.35 & 13.00 & 15.30 \\
  \hline
  4* & \thead{MuST-C original +\\Europarl +\\How2 synthetic +\\MuST-C synthetic} & 22.75 & 21.31 & 14.00 & 16.45 \\
  \hline
  5* & \thead{Finetune 3*\\on MuST-C original} & 23.60 &  22.26 & 13.71 & 15.30 \\
  \hline
  6* &  \thead{Finetune 3*\\on MuST-C original+\\MuST-C synthetic} & 23.64 & 22.23 & 13.67 & 15.29 \\
  \hline
  \hline
  \textbf{7} &  \textbf{\thead{Ensemble (1 to 6)}} & \textbf{25.22} & \textbf{23.80} & \textbf{15.20} & \textbf{16.53} \\
  \hline
 \end{tabular}
\caption{Detokenized case-sensitive BLEU scores  for  different experiments - * represents experiments that apply SpecAugment.}
\label{table:offlineresults}
\end{table*}

\begin{table}
 \centering
 \begin{tabular}{| c | c | c | c |} 
  \hline
  \textbf{\thead{Model}} & \textbf{\thead{iwslt seg}} &  \textbf{\thead{ASR seg}} \\ 
  \hline
  \hline
  3* & constrastive5 & constrastive3 \\ 
  \hline
  \hline
  4*  & constrastive4 & constrastive2 \\ 
  \hline
  \hline
  Ensemble & constrastive1 & \textbf{primary} \\
  \hline
 \end{tabular}
 \caption{The ranking of out submitted systems. Model 3* and 4* are respectively corresponding to No.3* and No.4* of Table \ref{table:offlineresults}.}
 \label{table:offlinesubmissionoverview}
\end{table}

\subsection{Architecture} 

We reuse our last year attention-based encoder-decoder architecture. As illustrated in Figure~\ref{fig:1}, the encoder has two VGG-like~\cite{simonyan2014very} CNN blocks followed by five stacked 1024-dimensional BLSTM layers.
Each VGG block is a stack of two 2D-convolution layers followed by a 2D-maxpooling layer aiming to reduce both time ($T$) and frequency ($D$) dimensions of the input speech features by a factor of $2$. After these two VGG blocks, input speech features' shape is transformed from $(T \times D)$ to $(T/4 \times D/4)$.
We used Bahdanau's attention mechanism~\cite{bahdanau2014neural} in all our experiments. The decoder is a stack of two LSTM layers 1024
dimensional memory cells. We would like to mention that Transformer based models have also been tested using the default ESPnet architecure and showed weaker results compared to the LSTM-based encoder-decoder architecture.

\mypar{Hyperparameters' details}
All of our models are trained in maximum 20 epochs, with early stopping after $3$ epochs if the accuracy on the development set does not improve. Dropout is set to $0.3$ on the encoder part, and Adadelta is chosen as our optimizer. During decoding time, the beam size is set to $10$. We prevent the models from generating too long sentences by setting a $maxlenratio\footnote{$maxlenratio=\dfrac{maximum\_output\_length}{encoder\_hidden\_state\_length}$}=1.0$. 
All our end-to-end models are similar in terms of architecture. They are different mainly in the following aspects:
(1) training corpus; (2) type of tokenization units;\footnote{All systems use 182 output caracter tokens except system 1 which has 201}  (3) fine-tuning and pretraining strategies. 
Description of different models and evaluation results are given in Section~\ref{sec:lessons}. 

\begin{table*}
 \centering
 \begin{tabular}{| c | c | c | c | c | c | c | c |} 
  \hline
  \textbf{No.} & \textbf{Set} & \textbf{BLEU} & \textbf{TER} & \textbf{BEER} & \textbf{CharacTER} & \textbf{BLEU(ci)} & \textbf{TER(ci)} \\ 
  \hline
  \hline
  1 & 2019.contrastive1 & 17.57 & 71.68 & 47.24 & 58.03 & 18.64 & 69.66 \\
  \hline
  2 & 2019.contrastive2 & 17.83 & 71.60 & 48.66 & 53.49 & 18.9 & 69.26 \\
  \hline
  3 & 2019.contrastive3 & 19.03 & 66.96 & 49.12 & 54.10 & 19.97 & 65.01 \\
  \hline
  4 & 2019.contrastive4 & 15.08 & 78.79 & 45.87 & 59.06 & 16.06 & 76.62 \\
  \hline
  5 & 2019.contrastive5 & 15.87 & 74.17 & 46.18 & 59.96 & 16.86 & 72.15 \\
  \hline
  \textbf{6} & \textbf{2019.primary} & \textbf{20.19} & \textbf{66.38} & \textbf{49.89} & \textbf{52.51} & \textbf{21.23} & \textbf{64.26} \\
  \hline
  \hline
  7 & 2020.contrastive1 & 18.47 & 71.85 & 48.92 & 55.83 & 19.46 & 69.88 \\
  \hline
  8 & 2020.contrastive2 & 19.31 & 69.30 & 49.55 & 52.68 & 20.36 & 67.14 \\
  \hline
  9 & 2020.contrastive3 & 20.51 & 64.88 & 50.19 & 53.06 & 21.5 & 62.99 \\
  \hline
  10 & 2020.contrastive4 & 15.48 & 83.45 & 46.68 & 57.56 & 16.42 & 81.33 \\
  \hline
  11 & 2020.contrastive5 & 16.5 & 75.15 & 47.23 & 57.90 & 17.42 & 73.22 \\
  \hline
  \textbf{12} & \textbf{2020.primary} & \textbf{22.12} & \textbf{63.87} & \textbf{51.20} & \textbf{51.46} & \textbf{23.25} & \textbf{61.85} \\
  \hline
 \end{tabular}
\caption{IWSLT 2020 official results (offline track) on  tst2019 and tst2020.}
\label{table:iwslt2020officialresults}
\end{table*}

\subsection{Speech segmentation}   
\label{sec:seg}
Two types of segmentation of evaluation and development data were used for experiments and submitted systems: segmentation provided by the IWSLT organizers and automatic segmentation based on the output of an  ASR system.

The ASR system, used to obtain automatic segmentation, was trained with the Kaldi speech recognition toolkit~\cite{Povey11thekaldi}. An acoustic model  was trained using the TED-LIUM~3 corpus~\cite{hernandez2018ted}.\footnote{The off-limit TED talks from IWSLT-2019  were excluded from the training subset}
This ASR system produces recognized words with timecodes (start time and duration for each word). Then we form the speech segments based on this output following the rules: (1) if  silence duration between two words is longer than a given threshold $\Theta=0.65$  seconds, we split the audio file; (2)~if the number of words in the current speech segment exceeds $40$, then $\Theta$ is reduced to $0.15$ seconds in order to avoid too long segments.
 These thresholds have been optimised to  get segment duration distribution in the development and evaluation data that is similar to the one observed in the training data. It will be shown in next subsection that this ASR segmentation improves results over the provided segmentation when the latter is noisy (see experimental results on iwslt/tst2015).

\subsection{Experiments and results} 
\label{sec:lessons}

After witnessing the benefit of merging different corpora from our submission last year~\cite{nguyen2019ontrac}, we continue exploring different combinations of corpora in this submission. As shown in the first two rows of Table~\ref{table:offlineresults}, merging How2 synthetic with the baseline (MuST-C original + Europarl) does not bring significant improvement. It is noticeable that this pool is worse than the baseline on both tst2015 (iwslt seg) and tst2015 (ASR seg). However, we find that applying data augmentation (SpecAugment) on this same combination helps outperform the baseline on every investigated testset, most significantly on MuST-C tst-COMMON, and MuST-C tst-HE. Therefore, SpecAugment is consistently applied to all the experiments that follow. Adding MuST-C synthetic to this pool surprisingly decreases BLEU scores on both MuST-C testsets, while significantly increases the scores on both tst2015 (iwslt seg) and tst2015 (ASR seg). Not being able to investigate further on this matter due to time constraint, instead of fine-tuning 4*, we decided to fine-tune 3*, which performs reasonably well among all the testsets, on MuST-C original and MuST-C original+synthetic. We witness that the impact of fine tuning is very limited. One can also see once again that adding MuST-C synthetic does not make much difference. Finally, the last row of the table shows the results of ensembling all six models at decoding time. It is clear from the table that ensembling yields the best BLEU scores across all the testsets.

\subsection{Overview of systems submitted}

Two conclusions that can be drawn from Table~\ref{table:offlineresults} are (1) ensembling all six models is the most promising among all presented models, (2) our own segmentation (tst2015 ASR segmentation) is better than the default one. Therefore, we choose as our primary submission the translations of the ASR segmentations generated by the ensemble of all six models. Model 3* and 4* (Table \ref{table:offlineresults}) are also used to translate our contrastive submission runs, whose ranks are shown in Table~\ref{table:offlinesubmissionoverview}. The official results for all our submitted systems can be found in Table \ref{table:iwslt2020officialresults}. They confirm that our segmentation approach proposed is beneficial.

%% file: online.tex
\section{Simultaneous Speech Translation Track}\label{sec:simultaneous}
In this section, we describe our submission to the Simultaneous Speech Translation (SST) track.
Our pipeline consists of an automatic speech recognition (ASR) system followed by an online machine translation (MT) system.
We first define our online ASR and MT models in \sect{online-asr} and \sect{online-mt} respectively. 
Then, we outline in \sect{asr-mt} how we arrange the two systems for the speech-to-text subtask. 
We detail our experimental setup and report our results on the text-to-text subtask in \sect{t2t} and on the speech-to-text in \sect{s2t}.

\subsection{Online ASR}\label{sec:online-asr}
Our ASR system is a hybrid HMM/DNN system trained with lattice-free MMI \cite{povey2016chain},
using the Kaldi speech recognition toolkit~\cite{Povey11thekaldi}. 
The acoustic model (AM) topology consists of a Time Delay Neural Network (TDNN) followed by a stack of 16 factorized TDNNs~\cite{povey2018tdnnf}.
The acoustic feature vector is a concatenation of 
40-dimensional MFCCs without cepstral truncation (MFCC-40) 
and 100-dimensional i-vectors for speaker adaptation \citep{Dehak10ivector}.
Audio samples were randomly perturbed in speed and amplitude during the training
process. This approach is commonly called audio augmentation and is known to
be beneficial for speech recognition \cite{ko2015perturb}. 

\mypar{Online decoding with Kaldi}
The online ASR system decodes under a set of rules to decide when to stop decoding and output a transcription. An endpoint is detected if either of the following conditions is satisfied: 
\begin{enumerate}[label=(\alph*), itemsep=0pt]
    \item After $t$ seconds of silence even if nothing was decoded.
    \item After $t$ seconds of silence after decoding something, if the final-state was reached with $\text{cost}_\text{relative} < c$.
    \item After $t$ seconds of silence after decoding something, even if no final-state was reached. 
    \item After the utterance is $t$ seconds long regardless of anything else.
\end{enumerate}
Each rule has an independent characteristic time $t$ and condition (b) can be duplicated with different times and thresholds $(t, c)$.
The value of $\text{cost}_\text{relative}$ reflects the quality of the output, it is \emph{null} if a final-state of the decoding graph had the best cost at the final frame, and \emph{infinite} if no final-state was active.

\subsection{Online MT}\label{sec:online-mt}
Our MT systems are Transformer-based \citep{Vaswani17nips} wait-k decoders with unidirectional encoders.
Wait-$k$ decoding starts by reading $k$ source tokens, then alternates between reading and writing a single token at a time, until the source is depleted, or the target generation is terminated. 
With a source-target pair $(\x,\y)$, the number of source tokens read when decoding $y_t$ following a \waitk policy is $\zw_t = \min(k+t-1, \lx).$ 
To stop leaking signal from future source tokens, the energies of the encoder-decoder multihead-attention are masked to only include the $z_t$ tokens read so far. 

Unlike Transformer \waitk models introduced in \citet{Ma19acl} where the source is processed with a bidirectional encoder, we opt for a unidirectional encoding of the source.
In fact, this change alleviates the cost of re-encoding the source sequence after each read operation. Contrary to offline task, where bidirectional encoders are superior, unidirectional encoder achieve better quality-lagging trade-offs in online MT.

\citet{Ma19acl} optimize their models with maximum likelihood estimation \wrt a single wait-$k$ decoding path $\Zw k$:
\begin{align}
\log p(\y\cond\x,\Zw k) = 
\sum_{t=1}^\ly \log p_\theta(y_t|\y_{<t},\x_{\leq \zw_t}).
\end{align}

Instead of optimizing a single decoding path, we jointly optimize across multiple wait-$k$ paths. 
The additional loss terms provide a richer training signal, and potentially yield models that could perform well under different lagging constraints. 
Formally, we consider an exhaustive set of wait-$k$ paths and in each training epoch we encode the source sequence then uniformly sample a path to decode with. As such, we optimize:
\begin{align}
    \Zm=\left\{\Zw k \cond k\in\{1,\ldots,\lx\}\right\},\\
   \E{\z}{\log p(\y|\x, \z)} {\approx}\sum_{\z\sim \in\Zm}\!\log p_\theta(\y|\x,\z).
\end{align}
We will refer to this training with \emph{multi-path}. 

\begin{algorithm}[t]
\caption{ASR+MT decoding algorithm}\label{alg:asr-mt}
\begin{algorithmic}
\algrenewcommand\algorithmicindent{2mm}
\State\algorithmicrequire source audio blocks $\x$.
\State\algorithmicensure translation hypothesis $\y$.
\State\textbf{Initialization:}~action=\ra, $z{=}0$, $t{=}1$, 
\State\phantom{\bf Initialization:}~$\xasr{=}()$, $\y{=}(\bos)$
\State\textbf{Hyper-parameters}~$sz, \alpha, \beta$.
\While{$y_t \neq \eos$}
\While{$\text{action} = \ra \land z<\lx$}
\State Read $sz$ elements from $\x$. $z\pluseq sz$
\State Feed the new audio blocks to the ASR system.
\If{Endpoint detected $\lor\,z=\lx$}
\State Output transcription and append it to $\xasr$.
\State action = \wa
\EndIf
\EndWhile 
\If{$\ly < \alpha|\xasr| + \beta$}
\State Given $\y$ and $\xasr$, predict the next token $y_{t+1}$ 
\State $t\pluseq 1$
\Else
\State action = \ra
\EndIf
\EndWhile 
\end{algorithmic}
\end{algorithm}

\subsection{Cascaded ASR+MT}\label{sec:asr-mt}

For speech-to-text online translation we pair an ASR system with our online MT system and decode following the algorithm described in \alg{asr-mt}.

In this setup, the lagging is controlled by the endpointing of the ASR system. The online MT system follows the lead of the ASR and translates prefix-to-prefix. Since the MT system is not trained to detect end of segments and can only halt the translation by emitting $\eos$, we constrain it to decode $\alpha |\xasr| + \beta$ tokens, where $\xasr$ is the partial transcription and $(\alpha, \beta)$ two hyper-parameters.

Along with the hyper-parameters of the ASR's endpointing rules, 
we tune $(\alpha, \beta)$ on a development set to achieve good latency-quality trade-offs.

\subsection{Text-to-text translation subtask}\label{sec:t2t}

\begin{table}
\centering
\resizebox{0.98\linewidth}{!}{%
\input{tables/mt_data}
}
\caption{Parallel training data for the MT systems.}
\label{tab:mt-data}
\end{table}

\mypar{Training MT}
We train our online MT systems on English-to-German MuST-C~\cite{mustc19} and WMT'19 data,\footnote{\url{http://www.statmt.org/wmt19/}} namely, Europarl~\cite{Koehn05europarl}, News Commentary~\cite{Tiedemann12news} and Common Crawl~\cite{Smith13acl}. We remove pairs with a length-ratio exceeding 1.3 from Common Crawl and pairs exceeding a length-ratio of 1.5 from the rest.
We develop on MuST-C dev and report results on MuST-C tst-COMMON.
For open-vocabulary translation, we use \emph{SentencePiece} \cite{Kudo18acl} to segment the bi-texts with byte pair encoding~\cite{sennrich2016neural}. This results in a joint vocabulary of 32K types. Details of the training data are provided in \tab{mt-data}.

We train Transformer \emph{big} architectures and tie the embeddings of the encoder with the decoder's input and output embeddings.
We optimize our models with label-smoothed maximum likelihood~\cite{Szegedy16cvpr} with a smoothing rate $\epsilon = 0.1$. The parameters are updated using Adam~\cite{Kingma15iclr} $(\beta_1,\beta_2 = 0.9,0.98$) with a learning rate that follows an inverse square-root schedule. We train for a total of 50K updates and evaluate with the check-pointed weights corresponding to the lowest (best) loss on the development set.
Our models are implemented with Fairseq \cite{ott2019fairseq}.
We generate translation hypotheses with greedy decoding and 
evaluate the latency-quality trade-off by measuring case-sensitive detokenized BLEU~\cite{papineni02} and word-level Average Lagging (AL)~\cite{Ma19acl}.

\begin{figure}[t]
\centering
\input{figures/tst_common_t2t.tex}
\caption{%
[Text-to-Text]
Latency-quality trade-offs evaluated on MuST-C tst-COMMON with greedy decoding. 
Offline systems have an AL of 18.55 words.
The red vertical bars correspond to the AL evaluation thresholds.
}\label{fig:t2t}
\end{figure}
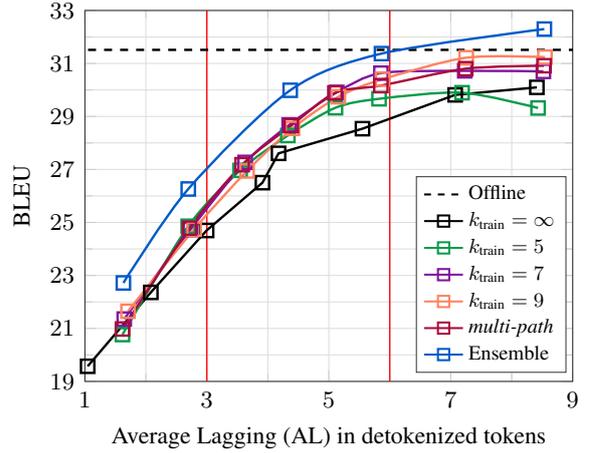

\mypar{Results}
We show in \fig{t2t} the performance of our systems on the test set (MuST tst-COMMON) measured with the provided evaluation server.\footnote{\url{https://github.com/pytorch/fairseq/blob/simulastsharedtask/examples/simultaneous_translation}} We denote with $\ktr{=}\infty$ a unidirectional model trained for wait-until-end decoding \ie reading the full source before writing the target. We evaluate four \waitk systems each trained with a value of $\ktr$ in $\{5,7,9,\infty\}$ and decoded with $\kev$ ranging from 2 to 11. We then ensemble the aforementioned \waitk models and evaluate a \emph{multi-path} model that jointly optimizes a large set of \waitk paths. 
The results demonstrate that \emph{multi-path} is competetive with \waitk without the need to select which path to optimize (some values of $k$, \eg 5, underperform in comparison).
Ensembling the \waitk models gives a boost of 1.43 BLEU points on average.

\subsection{Speech-to-text translation subtask}\label{sec:s2t}

\mypar{Training ASR}
We train our system following the \emph{tedlium} 
recipe\footnote{\url{https://github.com/kaldi-asr/kaldi/tree/master/egs/tedlium}} 
while adapting it for the IWSLT task. 
The TDNN layers have a hidden dimension of 1536 with a linear bottleneck dimension of 160 in the factorized layers.
The i-vector extractor is trained on all acoustic data (speech perturbed + speech) using a 10s window.
The acoustic training data includes TED-LIUM 3, How2 and Europarl. These corpora are detailed in \tab{asr-data} and represent about 900 hours of audio.

\begin{table}
\centering
\resizebox{0.98\linewidth}{!}{
    \begin{tabular}{lrrr}
    \toprule
    Corpus    &  \#hours & \#words & \#speakers \\
    \midrule
    TED-LIUM 3     &  452   & 5.05M & 2,028 \\
    How2 & 365 &  3.31M & 13,147\\
    Europarl & 94 & 0.75M & 171 \\
    \bottomrule
    \end{tabular}
}
\caption{Corpora used for the acoustic model.}
\label{tab:asr-data}
\end{table}

As a language model, we use the 4-grams \emph{small} model provided with TED-LIUM 3. The vocabulary size is 152K, with 1.2 million of 2-grams, 622K 3-grams and 70K 4-grams.

The final system is tuned on TED-LIUM 3 dev and tested with TED-LIUM 3 test and MuST-C tst-COMMON. Results are shown in \tab{kaldi-results}.

\begin{table}[t]
\centering
\resizebox{0.95\linewidth}{!}{%
    \begin{tabular}{llr}
    \toprule
    Decoding & Corpus & WER  \\
    \midrule
    Offline & TED-LIUM 3 dev & 7.65 \\
    Offline & TED-LIUM 3 test & 7.84 \\
    Offline & MuST-C tst-COMMON & 14.2 \\
    Online & MusT-C tst-COMMON & 16.3 \\
    \bottomrule 
\end{tabular}
}
\caption{WERs for the ASR system with offline and online decoding (AL=5s for online)
}
\label{tab:kaldi-results}
\end{table}

\mypar{Training MT}
To train the MT system for the ASR+MT cascade we process  source-side data (English) to match  transcriptions of the ASR. This consists of lower-casing, removing punctuation and converting numbers into letters.
For this task we use two distinct English and German vocabularies of 32K BPE tokens each.
We train Transformer \emph{big} architectures with tied input-output decoder embeddings following the setup described in \sect{t2t}.

\begin{figure}[t]
\centering
\input{figures/tst_common_s2t.tex}
\caption{%
[Speech-to-Text]
Latency-quality trade-offs evaluated on MuST-C tst-COMMON with greedy decoding. 
Offline systems have an AL of 5806 ms.
The red vertical bars correspond to the AL evaluation thresholds.
}\label{fig:s2t}
\end{figure}
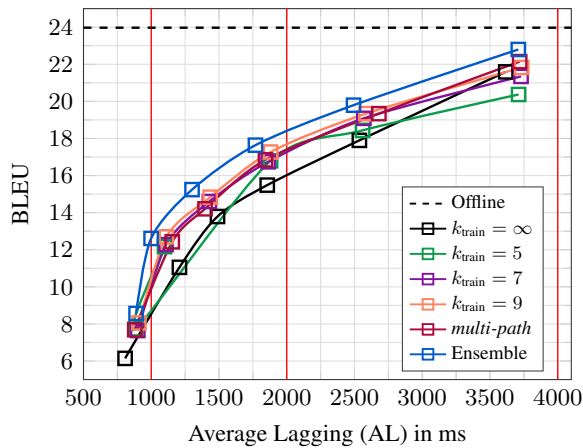

\mypar{Results}
Similar to the text-to-text subtask, we show our results in a plot of BLEU-to-AL in \fig{s2t}. The systems are evaluated on the test via the provided evaluation server where MuST-C's sentence-level aligned segments are streamed and decoded online and the lagging is measured in milliseconds.
Note that in this task we use a single ASR model and only ensemble the MT \waitk models. 
The cascade of an online ASR with \waitk MT follows the same trends as the text-to-text models.
In particular, \emph{multi-path} is competitive with specialized \waitk models and ensembling boosts the BLEU scores by 0.67 points on average.

%% file: tables/mt_data.tex
\begin{tabular}{lrrr}
\toprule
                     &    Pairs & \makecell[r]{English\\words} & \makecell[r]{German\\words}\\
\midrule
Europarl            &  1,730K &  43,7M  & 41,1M\\
Common Crawl        &  1,543K &  31,0M  & 30,0M\\
News Commentary     &    320K &   7,0M  &  7,2M\\
MuST-C               &    214K &   3,9M  &  3,7M \\
\bottomrule
\end{tabular}

%% file: figures/tst_common_t2t.tex
\begin{tikzpicture}
\pgfkeys{/pgf/number format/.cd,1000 sep={}}
\pgfmathsetmacro{\ALWue}{18.551}
\pgfmathsetmacro{\Wue}{31.513}


\begin{axis}[
    height=6.5cm, width=8.cm, 
    grid=both, y axis line style=-,
    legend style={
        font=\small, 
        legend cell align=left},
    ymin=19, ymax=33,
    ytick={19,21,...,35},
    minor y tick num=1,
    xmin=1,xmax=9,
    xtick={1,3,...,11},
    minor x tick num=1, 
    tick label style={font=\small},
    label style={font=\small},
    xlabel=Average Lagging (AL) in detokenized tokens,
    ylabel=BLEU,
    every axis plot/.append style={line width=0.9pt, mark size=2.5pt, mark=square},
    legend to name=t2t
   ]
\foreach \x in {3,6,15}
    \addplot [mark=none, line width=0.5pt, red, forget plot] coordinates {(\x, 19) (\x, 33)};

\addplot [mark=none, black, dashed] coordinates {(0, \Wue) (10, \Wue)};
\addlegendentry{Offline}

\addplot[black]
table [y=BLEU,x=AL]{results/tst_common_t2t_wue.dat};
\addlegendentry{$\ktr=\infty$}

\addplot[smooth, p1green]
table [y=BLEU,x=AL]{results/tst_common_t2t_wait5.dat};
\addlegendentry{$\ktr=5$}

\addplot[smooth, p1purple]
table [y=BLEU,x=AL]{results/tst_common_t2t_wait7.dat};
\addlegendentry{$\ktr=7$}

\addplot[smooth, p1orange]
table [y=BLEU,x=AL]{results/tst_common_t2t_wait9.dat};
\addlegendentry{$\ktr=9$}

\addplot[smooth, p1red]
table [y=BLEU,x=AL]{results/tst_common_t2t_multi.dat};
\addlegendentry{\emph{multi-path}}

\addplot[smooth, p1blue]
table [y=BLEU,x=AL]{results/tst_common_t2t_ensemble.dat};
\addlegendentry{Ensemble}

\end{axis}
\node[anchor=south east, scale=.85] at (rel axis cs: 0.88,0.0) {\pgfplotslegendfromname{t2t}};

\end{tikzpicture}

%% file: figures/tst_common_s2t.tex
\begin{tikzpicture}
\pgfkeys{/pgf/number format/.cd,1000 sep={}}
\pgfmathsetmacro{\ALWue}{5806}
\pgfmathsetmacro{\Wue}{23.972}


\begin{axis}[
    height=6.5cm, width=8.cm, 
    grid=both, y axis line style=-,
    legend style={
        font=\small, 
        legend cell align=left},
    xtick={0,500,...,4000},
    minor x tick num=1, 
    ytick={6,8,...,30},
    minor y tick num=1,
    tick label style={font=\small},
    label style={font=\small},
    xmin=500,xmax=4100,
    ymin=5, ymax=25,
    xlabel=Average Lagging (AL) in ms,
    ylabel=BLEU,
    every axis plot/.append style={line width=0.9pt, mark size=2.5pt, mark=square},
    legend to name=s2t
   ]

\foreach \x in {1000, 2000,4000}
    \addplot [mark=none, line width=0.5pt, red, forget plot] coordinates {(\x, 5) (\x, 26)};

\addplot [mark=none, black, dashed] coordinates {(500, \Wue) (4100, \Wue)};
\addlegendentry{Offline}

\addplot[smooth, black]
table [y=BLEU,x=AL]{results/tst_common_s2t_wue.dat};
\addlegendentry{$\ktr=\infty$}

\addplot[smooth, p1green]
table [y=BLEU,x=AL]{results/tst_common_s2t_wait5.dat};
\addlegendentry{$\ktr=5$}

\addplot[smooth, p1purple]
table [y=BLEU,x=AL]{results/tst_common_s2t_wait7.dat};
\addlegendentry{$\ktr=7$}

\addplot[smooth, p1orange]
table [y=BLEU,x=AL]{results/tst_common_s2t_wait9.dat};
\addlegendentry{$\ktr=9$}

\addplot[smooth, p1red]
table [y=BLEU,x=AL]{results/tst_common_s2t_multi.dat};
\addlegendentry{\emph{multi-path}}

\addplot[smooth, p1blue]
table [y=BLEU,x=AL]{results/tst_common_s2t_ensemble.dat};
\addlegendentry{Ensemble}

\end{axis}

\node[anchor=south east, scale=.8] at (rel axis cs: 0.96,-0.0) {\pgfplotslegendfromname{s2t}};

\end{tikzpicture}